\documentclass[10pt,twocolumn,letterpaper]{article}

\usepackage{iccv}
\usepackage{times}
\usepackage{epsfig}
\usepackage{graphicx}
\usepackage{amsmath}
\usepackage{amssymb}
\usepackage{balance}
\usepackage{booktabs}
\usepackage{enumitem}
\usepackage{mathtools}
\usepackage{makecell}
\usepackage{multirow}
\usepackage{pifont}
\usepackage{stackengine}
\newcommand\xrowht[2][0]{\addstackgap[.5\dimexpr#2\relax]{\vphantom{#1}}}
\usepackage[table]{xcolor}
\usepackage{soul}
\usepackage{array}
\newcolumntype{P}[1]{>{\centering\arraybackslash}p{#1}}
\usepackage[pagebackref=true,breaklinks=true,letterpaper=true,colorlinks,bookmarks=false]{hyperref}

\definecolor{g}{RGB}{34,139,34}  

\iccvfinalcopy % *** Uncomment this line for the final submission

\def\iccvPaperID{2705} % *** Enter the ICCV Paper ID here

\begin{document}

%%%%%%%%% TITLE
\title{ACE: Ally Complementary Experts for\\
Solving Long-Tailed Recognition in One-Shot}
%\title{Triple-branch Network for One-stage Long-tailed Recognition}

\author{Jiarui Cai, Yizhou Wang, Jenq-Neng Hwang\\
University of Washington\\
Seattle, WA, USA\\
{\tt\small \{jrcai, ywang26, hwang\}@uw.edu}
}

\maketitle
% Remove page # from the first page of camera-ready.
% \ificcvfinal\thispagestyle{empty}\fi

%%%%%%%%% ABSTRACT
\begin{abstract}
One-stage long-tailed recognition methods improve the overall performance in a ``seesaw'' manner, \ie, either sacrifice the head's accuracy for better tail classification or elevate the head's accuracy even higher but ignore the tail. Existing algorithms bypass such trade-off by a multi-stage training process: pre-training on imbalanced set and fine-tuning on balanced set. Though achieving promising performance, not only are they sensitive to the generalizability of the pre-trained model, but also not easily integrated into other computer vision tasks like detection and segmentation, where pre-training of classifiers solely is not applicable. In this paper, we propose a one-stage long-tailed recognition scheme,  ally complementary experts (ACE), where the expert is the most knowledgeable specialist in a sub-set that dominates its training, and is complementary to other experts in the less-seen categories without being disturbed by what it has never seen. We design a distribution-adaptive optimizer to adjust the learning pace of each expert to avoid over-fitting. Without special bells and whistles, the vanilla ACE outperforms the current one-stage SOTA method by 3$\sim$ 10$\%$ on CIFAR10-LT, CIFAR100-LT, ImageNet-LT and iNaturalist datasets. It is also shown to be the first one to break the ``seesaw'' trade-off by improving the accuracy of the majority and minority categories simultaneously in only one stage. Code and trained models are at \url{https://github.com/jrcai/ACE}. 
\end{abstract}

%%%%%%%%% BODY TEXT
\section{Introduction}

\begin{figure}[!t]
\begin{center}
\includegraphics[width=\linewidth]{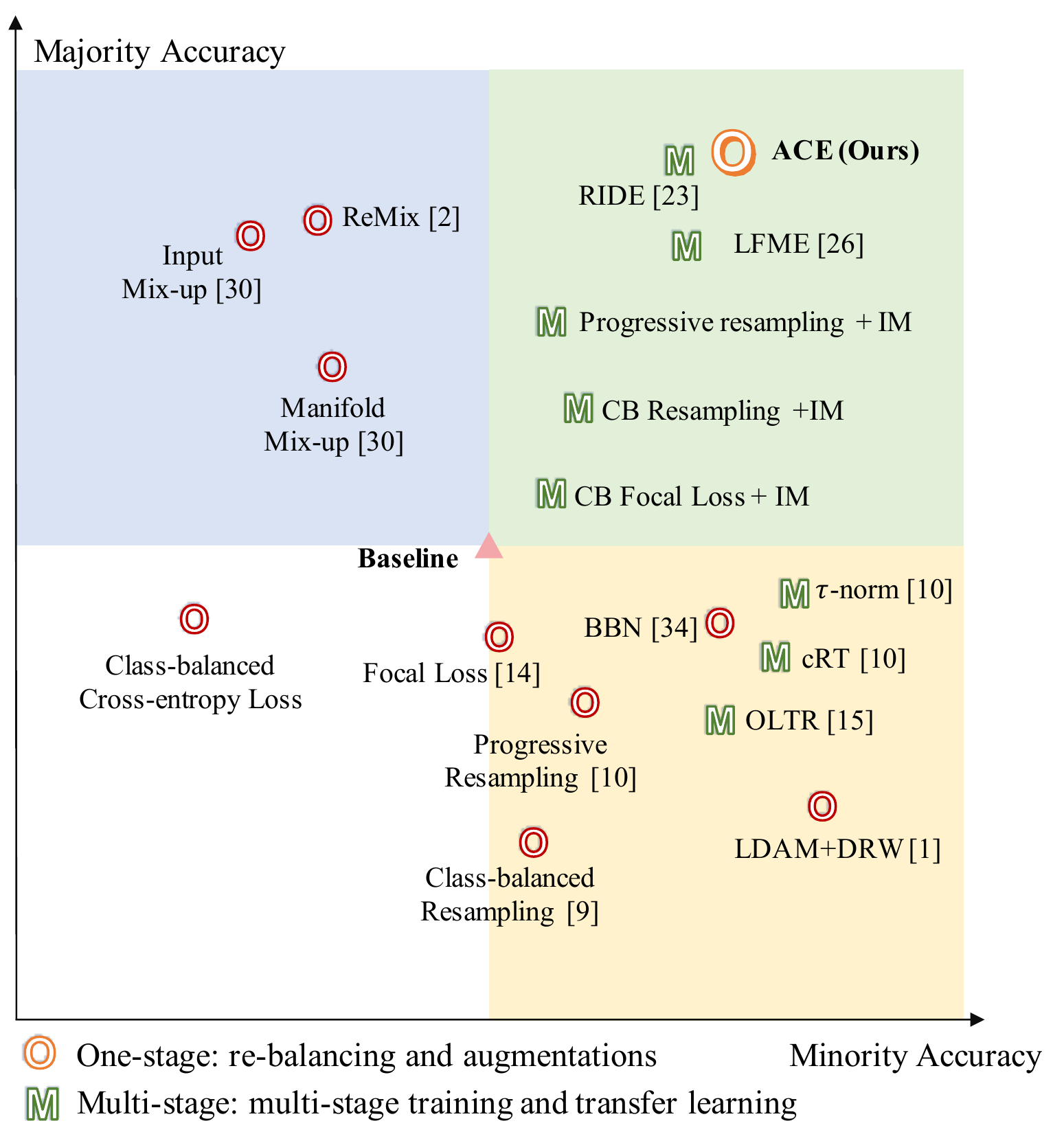}
\end{center}
   \caption{Performance of representative long-tailed recognition methods in terms of majority and minority classes compared to the baseline model (a ResNet). The results indicate most re-balancing methods improve the performance of minority categories by sacrificing that of the majority even with two-stage training (quadrant IV). Data augmentations are effective on the heads but slightly hurt the tails (quadrant II). The proposed ACE is the first one-stage SOTA method that improves the majority and minority simultaneously. Statistics for this figure are listed in the supplementary materials.}
\label{fig:related_works}
\end{figure}

% definition introduction
Object recognition is one of the most essential and substantial applications in computer vision. However, the performances of the state-of-the-art object recognition methods have limited capability on classifying real-world entities, which are skewed-distributed in a long-tailed manner naturally. Mostly driven by artificially-balanced datasets \cite{ deng2009imagenet,krizhevsky2009learning}, current models are dominated by the sample-rich classes and lose sight of the tails when adapting to long-tailed sets. Facing up to the reality, scarce as the tail categories are, they are of the same or even higher significance than the heads in various fields, such as biological species identification \cite{van2018inaturalist}, disease classification \cite{wang2020imbalance} and web-spam message detection \cite{zhao2020heterogeneous}. This long-lasting bottleneck significantly restricts classification-related computer vision tasks into practical use, including detection \cite{tan2020equalization, wu2020forest, yu2021towards} and instance segmentation \cite{wang2020devil, zang2021fasa}.

% overview of the related works

To ensure a well-accepted recognition capability over all categories, a tail-sensitive classifier becomes necessary. Existing solutions fall in three categories: one-stage \cite{Huang_2016_CVPR, wang2017learning}, two-stage with pre-training\cite{kang2019decoupling, cao2019learning}, and multi-stage multi-expert frameworks \cite{xiang2020learning, wang2020long}. The one-stage algorithms follow a straightforward idea to addressed the imbalance of training set by re-balancing, including re-sampling \cite{kang2019decoupling} and re-weighting \cite{cao2019learning, cui2019class, zhang2017range}. Despite the promotion of the tails, balancing techniques show an obvious ``seesaw'' phenomenon (Figure~\ref{fig:related_works}), that the accuracy of majority classes is sacrificed, indicating the under-representation of the heads. This raises a new concern that reducing the heads' accuracy might lead to more serious consequences. Taking the animal identification system as an example, some species are much richer in population than the endangered ones. Increasing the recognition accuracy of the snow leopards has little chance to be verified as they are rarely seen; on the contrary, failing to precisely classify two bird kinds can easily result in a misunderstanding of the local ecology. 

Literature in the recent years \cite{kang2019decoupling, wang2020long, xiang2020learning, zhang2021bag} handles the issue in a roundabout way: firstly train the feature extractor (backbone) with the whole imbalanced set for generalizable representation learning, then re-adjust the classifier by re-sampled data or build diverse experts for various tasks in cascading stages. Further improving the performance as they are, however, the general idea still holds old wine in a new bottle by making new trade-offs. To re-balance the data distribution, heavily relying on the well-adjusted pre-trained model and re-balancing skills make the frameworks sensitive to hyper-parameters and hard to find a sweet point. More importantly, the accumulated training steps make the multi-stage models redundant and less practical to be integrated with other tasks simultaneously, \eg, detection \cite{wang2020devil} and segmentation \cite{zang2021fasa}. To guarantee the plug-in and play property, it is thus highly desirable to have a classifier that overcomes the long-tail challenge with only one stage.

The hankerings of overcoming current long-tail challenges make us look more profoundly to the human intelligence. When human-beings make hard classification choices, saying diagnosis of diseases, it is advantageous to involve specialists’ insights who are well-aware of their own fields. Moreover, for the rare diseases, panel discussion and consultation are indispensable to exclude interfering potentials. Similarly, in the long-tailed issue, we are inspired to design a group of experts with \textit{complementary} skills: (1) they share elementary knowledge from the most diverse data source; (2) they are professional at splits of data respectively, and aware of what they do not specialize in; (3) opinions from the experienced experts (who see more data) are incorporated to complement the judgment from junior experts (who see less) for optimal decision.

Following the idea, we propose the Ally Complementary Experts (ACE) for one-stage long-tailed recognition. ACE is a multi-expert structure where experts are trained in parallel with a shared backbone. The experts are assigned with diverse but overlapping imbalanced subsets, to benefit from specialization in the dominating part. We also introduce a distribution-adaptive optimizer that controls the update of each expert according to the volume of its training set. Finally, the outputs of all experts are re-scaled and aggregated by data splits. ACE is trained end-to-end without any pre-training or staged-training.

We evaluate ACE on various widely-used long-tailed datasets, including CIFAR10-LT, CIFAR100-LT \cite{cui2019class}, ImageNet-LT \cite{liu2019large} and iNaturalist2018 \cite{van2018inaturalist} extensively with various experimental settings. Our method becomes the new SOTA among all one-stage long-tailed recognition methods with by 3-10$\%$ accuracy gain and is the first one that improves performance on all the three frequency groups (many-shot, medium-shot and few-shot). ACE also surpasses several multi-stage methods \cite{kang2019decoupling, kim2020m2m, liu2019large, xiang2020learning} by a large margin.

%------------------------------------------------------------------------
\section{Related Works}
\label{Related Works}
Methods for long-tailed recognition can be mainly grouped into three types: (1) readjustment of the data distribution; (2) two-stage training and transfer learning; (3) multi-expert/branch frameworks. 

\begin{table*}[t]
\small
\begin{center}
\begin{tabular}{c|c|c|c|c|c}
\toprule[1.5pt]
Method &  Data for Experts& Relationship of Experts & \makecell{Number of \\ Training Stages} & \makecell{Majority \\Gain}& \makecell{Minority \\ Gain} \\\hline\hline\xrowht[()]{6pt}
 LFME \cite{xiang2020learning} &  non-overlapping splits & independent & 2 & + & +\\ \hline\xrowht[()]{6pt}
 RIDE \cite{wang2020long}&  same full set & competing and complementary & 3 & ++ & +\\\hline\xrowht[()]{6pt}
\textbf{\textbf{ACE (Ours)}}& overlapping splits &  supportive and complementary & 1 & + & ++\\
\bottomrule[1.5pt]
\end{tabular}
\end{center}
\caption{Comparisons between the proposed method with two SOTA multi-expert networks.}
\label{tab:related_works}
\end{table*}

\subsection{On the Data: Re-balancing and Augmentations}
\label{sec:rebalancing}
Re-balancing consists of under-sampling of the head classes, over-sampling of the tail classes and re-weighting of the loss function by the frequency or importance of the samples \cite{japkowicz2002class, lin2017focal, cui2019class, cao2019learning}. Naive re-sampling in a class-balanced manner \cite{Huang_2016_CVPR,wang2017learning} can easily overfit on the sample-few classes, either constructing a less imbalanced distribution by square-root sampling \cite{mikolov2013distributed} or adjusting from instance-balanced to class-balanced sampling progressively \cite{cao2019learning, cui2019class, zhou2020bbn} is a more stable and promising alternative. 

Besides, strong data augmentations, which compensate for the insufficiency of data and improves the model's generalizability, could increase the diversity of the training set. Mixup \cite{zhang2017mixup} along with its long-tailed variant re-balanced Mix-up (ReMix) \cite{chou2020remix}; and tail classes synthesis \cite{zhang2021bag} are representative methods. However, the above algorithms commonly sacrifice the tails for the heads, or vice versa (Figure~\ref{fig:related_works}). The reason is the contradiction between representation learning and classifier learning, \ie, instanced-based (bias) sampling learns the most generalizable representations while the unbiased classifier is less likely to overfit the re-sampled set.

\subsection{On the Representation: Two-stage Training and Transfer Learning} 
The methods in the second category migrate the learned knowledge from the heads to tails by two-stage training or memory-based transfer learning. Deferred re-balancing by re-sampling (DRS) and re-weighting (DRW) scheme \cite{cao2019learning} train the classifier layers with re-balancing after obtaining good representation on the imbalanced set at the first stage. Kang \etal \cite{kang2019decoupling} propose $\tau$-norm and learnable weight scaling (LWS) to re-balance the decision boundaries of classifiers in the parameter domain. OLTR \cite{liu2019large} and inflated episodic memory (IEM) \cite{zhu2020inflated} utilize memory banks for prototype learning and knowledge transfer among classes. However, the use of re-balancing can still hurt the accuracy of heads, and the inevitable extra memory consumption potentially limits the deployment on large-scale datasets.

\subsection{Ensemble Methods: Multi-expert Networks} 
The recent trend on multi-expert or multi-branch networks shows the strong potential to address the long-tailed issue by treating the relatively balanced sub-groups separately. BBN \cite{zhou2020bbn}, which assigns two branches with normal and reversed sampling, respectively, incooperates a cumulative learning strategy to adjust the bilateral training. BBN merges the two-stage methods into one, but still suffers from the same drawbacks of slight degradation of head’s accuracy. LFME \cite{xiang2020learning} and RIDE \cite{wang2020long} are multi-expert architectures that learn diverse classifiers in parallel, combining with knowledge distillation and distribution-aware expert selection. The main difference between our proposal and these two state-of-the-art methods are summarized in Table~\ref{tab:related_works}. Though achieving impressive performance, both of them suffer from extensive hyper-parameter tuning to balance the multiple optimization functions. More importantly, the multi-stage training requirement makes them difficult to be integrated into other tasks, like detection and segmentation.

\section{Proposed Methodology}

\begin{figure*}[t]
\begin{center}
\includegraphics[width=\linewidth]{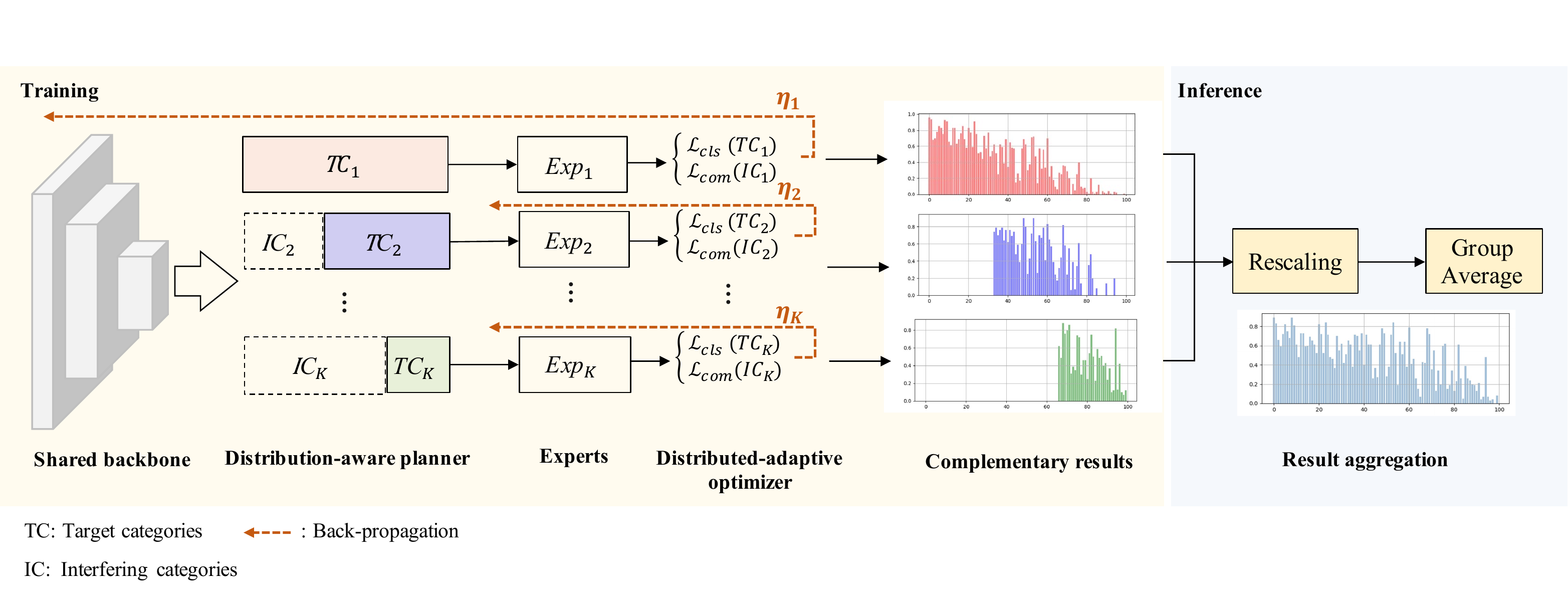}
\end{center}
   \caption{Network architecture of ACE. There are four components: (1) a shared backbone for representation learning; (2) a distribution-aware planner assigns diverse target categories (TC) and interfering categories to each expert, respectively; (3) a group of experts that learns to identify the TC with classification loss $L_{cls}$ and eliminate their effect on IC with complementary loss $L_{com}$; (4) a distribution-adaptive loss that adjust learning pace $\eta$ of each expert for simultaneous convergence. By allying complementary experts (ACE) in a group average manner, the aggregated prediction compromises the merits of all experts.}
\label{fig:ACE}
\end{figure*}

\subsection{The ACE Framework}
The architecture of the proposed Ally Complementary Experts (ACE) network is shown in Figure~\ref{fig:ACE}. 
Followed a shared backbone, multiple experts are branched out with individual learnable blocks and a prediction layer. A distribution-aware planner distributes diverse but overlapping category splits for each expert, including target categories (TC) and interfering categories (IC). These experts complement each other from three aspects: (1) the dominating categories in their TCs are different, so that the predictions have their own strengths; (2) the TCs are overlapping, especially on sample-few categories, thus the predictions support each other; (3) the experts learn to suppress the output of IC so that they will not bring ambiguity in the categories that have never been seen. To further accommodate the disparity in data, a distribution-adaptive optimizer is designed to guide the experts to update at their own paces. We use classification loss $L_{cls}$ and complement loss $L_{com}$ to train the model end-to-end in only one-shot. Finally, the predictions from the experts are aggregated by averaging over the re-scaled logits in each data split.

\subsection{Distribution-aware Planner}
The experimental fact that classifiers tend to have better performance on the majority categories than the minority on an imbalanced set is considered a drawback and avoided by existing methods. However, if each split’s prediction is obtained from a classifier that biases on it, we could expect accuracy gains everywhere. Therefore, we design a distribution-aware planner to assign each expert with a subset of the training set, which is also imbalanced and dominated by different splits, respectively. Formally, the process is as follows,

Given a training set $\mathbb{D}=\{X;Y\}$, where $X$ denotes the data and $Y$ denotes the corresponding class labels, with $C$ categories in total, for $K$ experts $\mathbb{E}=\{\mathcal{E}_1, \mathcal{E}_2, ..., \mathcal{E}_K\}$. Each $\mathcal{E}_i$ is assigned subset categories $\mathcal{C}_i$, where $C$ is assumed to be a multiple of $K$ for simplicity of discussion and without loss of generality, $|\mathcal{C}_1 \cup \mathcal{C}_2 \cup ... \cup \mathcal{C}_K|=C$ and $\mathcal{C}_i \cap \mathcal{C}_j \neq \varnothing$. 

Similar to the spirit of re-balancing, sample-few categories should be more exposed. Therefore, the $i$-th expert $\mathcal{E}_i$ is assigned target $\mathcal{C}_i $ and interfering classes $\widetilde {\mathcal{C}_i}$
\begin{equation}
     \begin{aligned}
        \mathcal{C}_i &=\{\frac{C}{K}(i-1)+1,\frac{C}{K}(i-1)+2, \frac{C}{K}(i-1)+3, ..., C \}, \\ 
        \widetilde {\mathcal{C}_i} &=\{1, 2, ...,\frac{C}{K}(i-1)\}.   
    \end{aligned}
\end{equation}

For a randomly-sampled mini-batch of training data $\mathcal{B} \subset \mathbb{D}$, $\mathcal{E}_i$ uses the corresponding sub-batch $\mathcal{B}_i=\{(x,y): (x,y) \in \mathcal{B}, y \in \mathcal{C}_i\}$. In this case, there is always an expert be presented for training with all the samples, and the smaller the class, the more experts are presented. Besides, with this data split mechanism, the medium-shot or few-shot classes have chances to dominate an expert, thus eliminating the bias towards the sample-rich classes. The network degenerates to a plain classifier if $K=1$. 

Similar to the existing methods, we use the ResNet \cite{he2016deep} as our backbone. The last residual block is duplicated for each expert, and followed by a learnable weight scaling (LWS) classifier \cite{kang2019decoupling}. The output logits (before SoftMax) of $\mathcal{E}_i$ is $\textbf{z}^{i} \in \mathbb{R}^{1\times|C|}$, which are fyrther adjusted to be $\hat{\textbf{z}_i}$ by the norm of the fully-connected layers' weights to have comparable scales: 
\begin{equation}
     \hat{\textbf{z}_i} = \frac{\|\textbf{w}_i\|^2}{\|\textbf{w}_1\|^2} \cdot \textbf{z}_i 
\end{equation}

The set of experts that trained with class $c$ is $\mathcal{S}^c$, then the output logit of class $c$ is the average among the outputs from $\mathcal{S}^c$, \ie, 
\begin{equation}
     \textbf{o}^c = \frac{1}{|\mathcal{S}_c|} \sum_{\mathcal{E}_i \in \mathcal{S}^c}\hat{\textbf{z}_i}
\end{equation}

SoftMax operation is applied on $\textbf{o}$ to obtain the classification confidence.

\subsection{Objective Functions}
Loss functions are applied on each expert separately instead of on the aggregated output $\textbf{o}$ to avoid a mixture of expect-specific features. We use the cross-entropy loss as the classification loss, with the sub-batch $\mathcal{B}_i$ for $\mathcal{E}_i$,
\begin{equation}
    L_{cls}^i(\mathcal{B}_i)=-\sum ^{\mathcal{C}_i}y\log(\sigma(\textbf{z}_i)), 
\end{equation}
where $\sigma (\cdot)$ represents the SoftMax operation.

In addition to classifying the assigned targeted class, each expertt’s response should not affect the other experts on the classes they have never seen, \ie, the interfering categories (IC). For the experts themselves, categories in IC are the main source of confusion as well. By eliminating the effect of IC, the experts work in a complementary manner rather than competitive. Hence, a regularization term to suppress the output of IC is necessary. We define the complement loss $L_{com}$ as

\begin{equation}
    L_{com}^i(\mathcal{B}_i)=\sum ^{\mathcal{C}}_{c_j \in \widetilde{\mathcal{C}_i}} \|\textbf{z}_i^{c_j}\|^2.
\end{equation}

The complement loss, which serves as a soft regularization in the optimization process, minimizes the logits of non-target categories for $\mathcal{E}_i$ so as to put down their effect. Detailed study of the impact of incorporating $L_{com}$ could be found in Sec~\ref{abs}.

Overall, the loss function for $\mathcal{E}_i$ is
\begin{equation}
    L_{\mathcal{E}_i}(\mathcal{B}_i)= L_{cls}^i + L_{com}^i.
\end{equation}

\subsection{Distributed-adaptive Optimizer}
Recall the Linear Scaling Rule \cite{goyal2017accurate} for training networks in mini-batches with stochastic gradient descent (SGD) optimizer: \textit{when the minibatch size is multiplied by k, multiply the learning rate by k. All other hyper-parameters (weight decay, momentum, \etc) are kept unchanged}. 

By this rule, to avoid over-fitting, the optimizer should be distribution-aware to assign smaller weights to $\mathcal{E}_i$ which is trained with less data. Denoted the base learning rate as $\eta_0$, which is the learning rate for the expert presented with all categories, the $i$-th expert is trained by,
\begin{equation}
    \eta_i = \eta_0 \cdot \frac{\sum_{c \in \mathcal{C}_i} {n_c}}{\sum^{\mathcal{C}} {n_j}},
\end{equation}
where $\mathcal{N}=\{n_1, n_2,... n_{C}\}$ are the number of samples in each class and $\mathcal{N}$ is assumed in the descending order. 

The loss of $\mathcal{E}_1$ updates the backbone and parameters of $\mathcal{E}_1$, and $L_{i}$ that $i>1$ only updates the expert itself. The reason is the errors likely duplicates because of data overlapping, which means the backbone could be corrected multiple times due to the same error. This is similar to the idea of re-weighting methods, as introduced in Section~\ref{sec:rebalancing} that hurt the representation learning. Therefore, only $\mathcal{E}_1$ updates the backbone.

\section{Experiments}
\subsection{Datasets and Protocols}
Generally, in long-tail recognition tasks, the classes are categorized into many (with more than 100 training samples), medium (with 20 $\sim$100 samples) and few (with less than 20 samples) splits \cite{liu2019large}. The \textit{imbalance factors}(IFs) of the long-tailed datasets, defined as the frequency of the largest class divided by the smallest class, vary from 10 to over 500 \cite{cui2019class, liu2019large, van2018inaturalist}.

\textbf{CIFAR100-LT and CIFAR10-LT} \cite{cui2019class} are artificially created from the balanced CIFAR dataset \cite{krizhevsky2009learning} by reducing training samples according to an exponential function $n = n_i \mu^i$, where $i$ is the class index, $n_i$ is the original number of samples and $\mu \in (0,1)$. We experiment with two commonly used IFs, 100 and 50. There are approximately 10K$\sim$13K training images and 10K testing images for each split. ResNet-32 is used as the base network, where the last residual block is tripled for the branches to be comparable with other methods. Following \cite{he2016deep}, for training samples, 4 pixels are padded on each side, following by a $32 \times 32$ random crop on the padded image or its horizontal flip. The network is trained by the stochastic gradient descent (SGD) optimizer with a momentum 0.9 for 400 epochs. The base learning rate is 0.1 and decreases by 0.1 at epoch 320 and 360, respectively. Mixup \cite{zhang2017mixup} augmentation is used with $\alpha$ as 0.3, and the last 20 epochs are trained without Mixup.

 \textbf{ImageNet-LT} \cite{liu2019large} is sampled from ImageNet-2012 \cite{deng2009imagenet} following the Pareto distribution with the power value $\alpha=6$. ImageNet-LT contains 115.8K images for 1000 categories, with a maximum of 1280 images per class and a minimum of 5 images per class. Following \cite{liu2019large, kang2019decoupling, xiang2020learning, zhang2021bag}, we use ResNet-10 as the backbone. To be comparable with \cite{kang2019decoupling,wang2020long}, we also report our results with ResNet-50 and ResNeXt-50 \cite{xie2017aggregated}. For data pre-processing, the training samples are resized to $256 \times 256$, then randomly cropped to $224 \times 224$ and flipped horizontally with a probability of 0.5; on testing, the aspect ratio of the testing sample is kept by first resizing proportionally its shorter side to 256 then crop $224 \times 224$ in the center. The networks are trained by the SGD optimizer with momentum 0.9 for 100 epochs. The base learning rate is 0.1 and decreases by 0.1 at epoch 120 and 160. Mixup augmentation is used with $\alpha$ as 0.3, and the last 20 epochs are trained without Mixup.
 
\textbf{iNaturalist2018} \cite{van2018inaturalist} is a real-world large-scale dataset for species identification of animals and plants. Following the literature, we use the 2018 version which contains 438K images for over 8K categories, with extremely imbalanced distribution (IF=512) and challenging fine-grained issues. We use ResNet-50 as the backbone, and the same prepossessing and training protocol as ImageNet-LT. Mixup augmentation is used with $\alpha$ as 0.3, and the last 20 epochs are trained without Mixup.

\subsection{Performance}
\textbf{Competing methods.} Generally, there are two types of the competing methods by whether or not there is a backbone pre-training stage. For one-stage type of methods, re-balancing of the long-tailed dataset is either by resampling, (\eg, class-balanced and progressively-balanced \cite{kang2019decoupling}), or reweighting (\eg, focal loss \cite{lin2017focal}, class-balanced focal loss \cite{cui2019class}, and LDAM \cite{cao2019learning}). Besides, strong augmentation tricks (\eg, mixup \cite{zhang2017mixup}, re-balanced mixup \cite{chou2020remix}, tail sample synthesis using class activation maps (CAM) \cite{zhang2021bag})) can also benefit the overall accuracy, especially the heads. Moreover, transfer learning in either image domain (major-to-minor translation \cite{kim2020m2m}) and in feature domain (OLTR \cite{liu2019large}) are proved useful. Logit Adjustment \cite{menon2020long} encourages a large relative margin between logits of rare
versus dominant labels with a one-stage loss. BBN \cite{zhou2020bbn} uses a two-branch architecture to combine normal sampling and distribution-reversed sampling progressively, improving the tail's accuracy in a large margin. The other type is two-stage methods. In the second stage, $\tau$-norm, LWS and cRT \cite{kang2019decoupling} retrain or fine-tune the classifier with a balanced dataset or unbiased classifier weights. LFME \cite{xiang2020learning} and RIDE \cite{wang2020long} are multi-branch assembled architectures with knowledge distillation. LFME uses a teacher-student network to train experts on many-/medium-/few-shot splits, while RIDE does not fix the number of branches and uses KL-divergence loss to force them to be experts on different groups. \\ 

\begin{table*}[t]
\small
\begin{center}
\begin{tabular}{c|P{3.5cm}|c|P{1.9cm}|P{1.9cm}|P{1.9cm}|P{1.9cm}}
\toprule[1.5pt]
 \multirow{2}{*}{\makecell[c]{\textbf{Type}}} & \multirow{2}{*}{\textbf{Method}} & \multirow{2}{*}{\makecell{\textbf{Multi-experts}}} & \multicolumn{4}{c}{\textbf{Accuracy}}\\
 \cline{4-7}
 & & & \textbf{All} & \textbf{Many} & \textbf{Medium} & \textbf{Few} \\\hline
\multirow{9}{*}{\makecell[c]{\textbf{One-Stage}}}
 & Baseline (ResNet-32) & & 38.3 & 65.2 & 37.1 & 9.1\\
 & CB resampling \cite{japkowicz2002class}\S & & 36.0 \small\textcolor{red}{(-1.7)} &59.0 \small\textcolor{red}{(-6.2)}& 35.4 \small\textcolor{red}{(-1.7)} &10.9 \small\textcolor{g}{(+1.8)} \\
 & Focal loss \cite{lin2017focal} & & 37.4 \small\textcolor{red}{(-0.9)} &64.3 \small\textcolor{red}{(-0.9)} & 37.4 \small\textcolor{g}{(+0.3)} &7.1 \small\textcolor{red}{(-2.0)} \\
 & CB Focal loss \cite{cui2019class}\S & & 38.7 \small\textcolor{g}{(+0.4)}& 65.0 \small\textcolor{red}{(-0.2)}& 37.6 \small\textcolor{g}{(+0.5)} &10.3 \small\textcolor{g}{(+1.2)} \\
 & Progressive \cite{kang2019decoupling}  & & 39.4 \small\textcolor{g}{(+1.1)}&63.3  \small\textcolor{red}{(-1.9)}& 38.8  \small\textcolor{g}{(+1.7)}&13.1  \small\textcolor{g}{(+4.0)}\\
 & ReMix \cite{chou2020remix} &  & 40.9 \small\textcolor{g}{(+2.6)}&69.6  \small\textcolor{g}{(+4.4)}& 40.7 \small\textcolor{g}{(+3.0)}& 8.8  \small\textcolor{red}{(-0.3)}\\
 & Mixup \cite{zhang2017mixup}  & &41.2 \small\textcolor{g}{(+2.9)}&70.7 \small\textcolor{g}{(+5.5)} & 40.4 \small\textcolor{g}{(+3.3)} & 8.8 \small\textcolor{red}{(-0.3)}\\
 & BBN \cite{zhou2020bbn}  &\checkmark & 39.4 \small\textcolor{g}{(+1.1)} &47.2 \small\textcolor{red}{(-18.0)} &49.4 \small\textcolor{g}{(+12.3)} &19.8 \small\textcolor{g}{(+10.7)} \\
 & Logit Adjustment \cite{menon2020long} & & 43.9 \small\textcolor{g}{(+5.6)} & - & - & -\\
 \rowcolor{yellow!20} \cellcolor{white} & \textbf{ACE (3 experts)}  & \checkmark & \textbf{49.4} \small\textcolor{g}{(+11.1)} &\textbf{66.1} \small\textcolor{g}{(+0.9)} & \textbf{55.7} \small\textcolor{g}{(+18.7)} &\textbf{23.5} \small\textcolor{g}{(+14.4)}\\
  \rowcolor{yellow!20} \cellcolor{white} & \textbf{ACE (4 experts)}  & \checkmark &\textbf{49.6} \small\textcolor{g}{(+11.2)} &\textbf{66.3} \small\textcolor{g}{(+1.1)} & \textbf{52.8} \small\textcolor{g}{(+15.7)} &\textbf{27.2} \small\textcolor{g}{(+18.1)}\\\hline\hline
\multirow{9}{*}{\makecell[c]{\textbf{Multi-Stage}}}&
 $\tau$-norm \cite{kang2019decoupling} & & 43.2 &65.7 & 43.6 &17.3 \\
 & cRT \cite{kang2019decoupling}  & & 43.3 &64.0 & 44.8 &18.1 \\
 & LDAM+DRW \cite{cao2019learning} & & 42.0 &61.5& 41.7 & 20.2 \\
 & LDAM+LFME \cite{xiang2020learning}  & \checkmark & 43.8 &- & - &- \\
 & LDAM+M2m \cite{kim2020m2m} & & 43.5 &- & - &- \\
 & CAM \cite{zhang2021bag}  &\checkmark & 47.8 &- & - &- \\
 & RIDE \cite{wang2020long} (2 experts)&\checkmark & 47.0 &67.9 & 48.4 & 21.8 \\
 & RIDE \cite{wang2020long} (3 experts)&\checkmark & 48.0 &68.1 & 49.2 & 23.9 \\
 & RIDE \cite{wang2020long} (4 experts)&\checkmark & 49.1 &69.3 & 49.3 & 26.0 \\
\bottomrule[1.5pt]
\end{tabular}
\end{center}
\caption{Top-1 accuracy on CIFAR100-LT-100. ($\cdot$) shows comparison to the baseline, where \textcolor{g}{increase} and \textcolor{red}{decrease} are represented by color. Our ACE is the only one-stage method with performance gain on all groups and of the best over all categories. \S: CB represents class-balanced.}
\label{tab:performance_cifar100}
\end{table*}

\begin{table}
\small
\begin{center}
%\begin{tabular}{c|c|P{2.1cm}|P{2.1cm}|P{2.1cm}|P{2.5cm}}
\begin{tabular}{c|c|c|c|c}
\toprule[1.5pt]
\multirow{2}{*}{\textbf{Method}} & \multicolumn{3}{c|}{\textbf{ImageNet-LT}} & \textbf{iNaturalist}\\
 \cline{2-5}
 & Res10 & Res50 & ResX50 & Res50\\\hline
 Baseline &   20.9 & 41.6 & 44.4 & 66.1 \\
 FSLwF \cite{gidaris2018dynamic} & 28.4 &- &- &- \\
 Range Loss \cite{zhang2017range} & 30.7 &- &- &- \\
 Lifted Loss \cite{oh2016deep} & 30.8 &- &- &- \\
 Focal loss \cite{lin2017focal} & 30.5&- &- &60.3 \\
 CB Focal loss \cite{cui2019class}  &- &- &- &61.1 \\
 BBN \cite{zhou2020bbn}  &- &48.3 &49.3 &68.0 \\
Logit Adj.\cite{menon2020long} & - & 51.1 & - & 66.4\\
\rowcolor{yellow!20} \textbf{ACE (3 experts)}  &\textbf{44.0} &\textbf{54.7} &\textbf{56.6} & \textbf{72.9} \\\hline
 OLTR \cite{liu2019large}  &34.1 &- &46.3 &63.9 \\
 NCM \cite{kang2019decoupling} & 35.5 & 44.3 & 47.3 & - \\
 LDAM+DRW \cite{cao2019learning}  &36.0 &- &- &68.0 \\
 cRT \cite{kang2019decoupling} & 41.8 & 47.3 & 49.5 & 65.2 \\
 $\tau$-norm \cite{kang2019decoupling} & 40.6 & 46.7 & 49.4 & 65.6 \\
 LWS \cite{kang2019decoupling} & 41.4 & 47.7 & 49.9 & 65.9 \\
 CAM \cite{zhang2021bag} & 43.1 & - & - & 70.9 \\
 LFME \cite{xiang2020learning} & 38.8 & - & - & - \\
 RIDE \cite{wang2020long}\dag& -& 54.4 & 55.9 & 71.4\\
 RIDE \cite{wang2020long}\ddag& -& 54.9 & 56.4 & 72.2\\
%& RIDE \cite{wang2020long} (4 experts) & -& 55.4 & 56.8 & 72.6\\
\bottomrule[1.5pt]
\end{tabular}
\end{center}
\caption{Top-1 accuracy on ImageNetLT and iNaturalist2018. Detailed results on each group are listed in the supplementary materials. Overall, it shows the multi-expert/branch architecture outperforms the re-balancing methods. Our ACE has consistent performance gain comparing with other one-stage methods with multiple backbones, and is comparable with multi-stage methods.\dag:2 experts, \ddag:3 experts.}
\label{tab:performance_imagenet}
\end{table}

\begin{figure}[h]
\begin{center}
\includegraphics[width=0.95\linewidth]{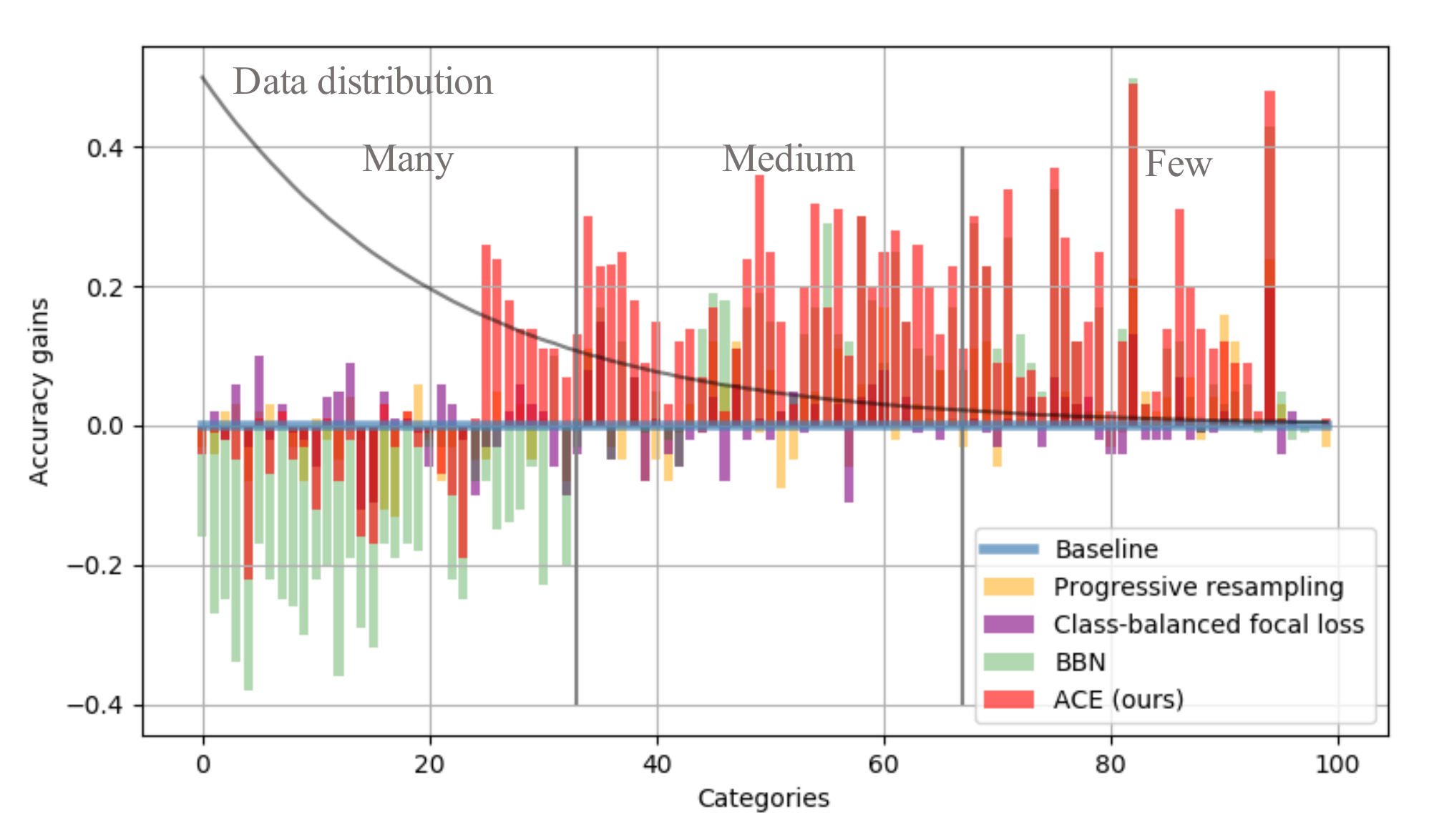}
\end{center}
   \caption{Accuracy gain comparisons between representative one-stage long-tailed recognition methods and baseline. While other methods decrease the majority accuracy, our ACE improves the many, medium and few groups all together.}
\label{fig:result}
\end{figure}

\textbf{CIFAR-LT} Table~\ref{tab:performance_cifar100} shows the proposed ACE performs the best among all one-stage methods and surpasses other multi-stage methods on CIFAR100-LT-100. We outperform the previous one-stage SOTA BBN by 9.0$\%$. Class-wise accuracy gain comparison with representative one-stage long-tailed recognition methods is shown in Figure~\ref{fig:result}. ACE has significant advantages in medium and few-shot categories. It is also the only method that improves all the groups by a single stage. Table~\ref{tab:performance_cifar} shows the top-1 accuracy on CIFAR10-LT and CIFAR100-LT with imbalance factor 50 and 100.\\

\begin{table}[t]
\small
\begin{center}
\begin{tabular}{p{2.7cm}|c c|c c}
\toprule[1.5pt]
 \multirow{2}{*}{\textbf{Method}} & \multicolumn{2}{c|}{\textbf{CIFAR100-LT}}  & \multicolumn{2}{c}{\textbf{CIFAR10-LT}}\\
 \cline{2-5}
 & \textbf{100} & \textbf{50} & \textbf{100} & \textbf{50} \\\hline
 Baseline & 38.3 & 42.1 & 69.8 & 75.2 \\
 Focal loss \cite{lin2017focal} & 37.4 & 42.4  & 70.4 & 75.3\\
 Mixup \cite{zhang2017mixup} & 39.5 & 45.0 & 73.1 & 77.8\\
 CB Focal loss \cite{cui2019class} & 38.7 & 46.2  & 74.6 & 79.3\\
 BBN \cite{zhou2020bbn} &39.4 &47.0  &79.8 &82.2\\
 Logit Adj.\cite{menon2020long} & 43.9 & - & 77.7 & -\\
 \rowcolor{yellow!20} \textbf{ACE (3 experts)} &\textbf{49.4} & \textbf{50.7} &\textbf{81.2} &\textbf{84.3} \\
  \rowcolor{yellow!20} \textbf{ACE (4 experts)} &\textbf{49.6} & \textbf{51.9} &\textbf{81.4} &\textbf{84.9} \\\hline
 LDAM+DRW \cite{cao2019learning} &42.0 &45.1 &77.0 &79.3\\
 LFME \cite{xiang2020learning}& 42.3 &- &- & -\\
 LDAM+M2m \cite{kim2020m2m}&43.5&- &79.1& - \\
 CAM \cite{zhang2021bag}&47.8 &51.7  &80.0 &83.6 \\
 RIDE \cite{wang2020long}&49.1 &- &- &- \\
\bottomrule[1.5pt]
\end{tabular}
\end{center}
\caption{Top-1 accuracy on CIFAR100-LT and CIFAR10-LT with imbalance factor 100 and 50.}
\label{tab:performance_cifar}
\end{table}

\textbf{ImageNet-LT and iNaturalist} We also report our performance on ImageNet-LT with various backbone models of ResNet-10, ResNet-50 and ResNeXt-50 as well as iNaturalist-LT with ResNet-50, shown in Table~\ref{tab:performance_imagenet}. Our method outperforms the BBN by 6.4$\%$ (ResNet-50) and  7.3$\%$ (ResNeXt-50) on ImageNet-LT and 3.9$\%$ on iNaturalist2018, respectively.

\subsection{How ACE Works}
\label{abs}
\begin{figure*}[t]
\begin{center}
\includegraphics[width=0.95\linewidth]{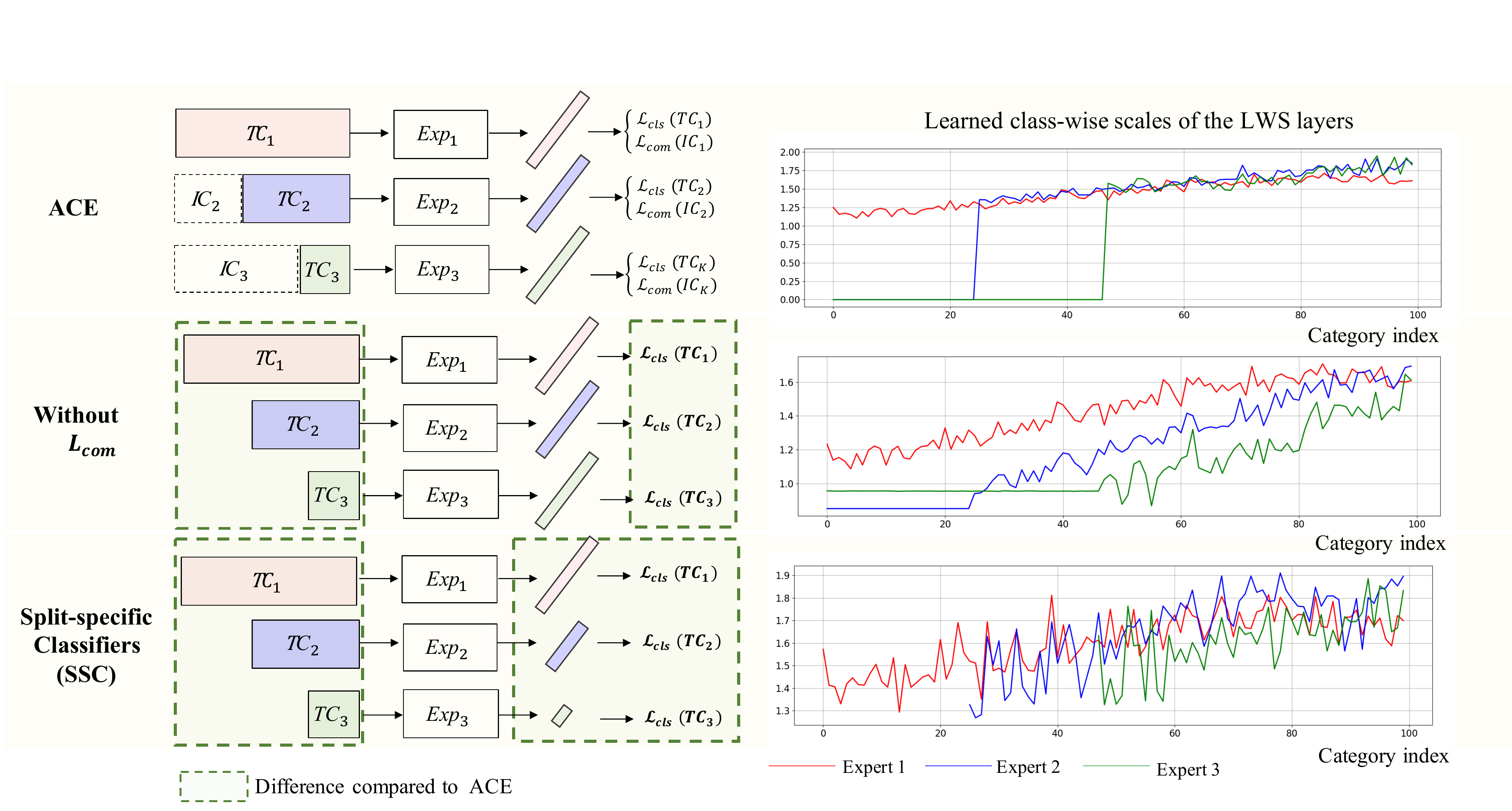}
\end{center}
   \caption{The design and classifier weights/bias of three model: ACE trained with complementary loss (top), ACE trained without complementary loss (middle), and split-specific classifiers (SSC) (bottom) trained on CIFAR100-LT-100. $\mathcal{E}_1$, $\mathcal{E}_2$ and $\mathcal{E}_3$ are plotted in red, blue and green colors, respectively. Complementary loss allows the experts work jointly in their common splits. Without the complementary loss, the experts trained with full batch has the largest scales on all splits and competes with the real dominating experts.}
\label{fig:abl-comloss}
\end{figure*}

\textbf{Complementary of experts.} We compare ACE with its two variants to show the effectiveness of its architecture, learning process and the loss function: one is training without $L_{com}$ and the other is the non-complementary architecture, which is called split-specific classifier (SSC). In the latter one, output dimensions of the classifier of $\mathcal{E}_i$ are the same as $|\mathcal{C}_i|$, \ie, $\textbf{z}^{i} \in \mathbb{R}^{1\times|\mathcal{C}_i|}$. In other words, the weights of non-target groups are set to be zero as a hard constraint, instead of learning to suppress them with $L_{com}$ in a soft regularization manner. The three architectures are shown in Figure~\ref{fig:abl-comloss} and their results are in Table ~\ref{tab:abs-lcom}. Figure~\ref{fig:abl-comloss} shows ACE with $L_{com}$ in the top row, where $\mathcal{E}_i$ learns similar scales over all the data splits and the scales of interfering classes are zeros. Therefore, all trained experts have an approximately equal contribution to the shared splits. We also observe that on the minority splits $j$ of $\mathcal{E}_i$  generates supportive results for $\mathcal{E}_j$ (\eg, $\mathcal{E}_1$ is peripheral in the few-shot split, and its scales are smaller than those of $\mathcal{E}_3$, so it is just a supplementary to $\mathcal{E}_3$'s output.). As seen from the middle row of Figure \ref{fig:abl-comloss}, by splitting the data to complementary batches, but without $L_{com}$, all experts  compete with one anoother in the common splits. For example, $\mathcal{E}_1$ is strong over all categories though it is less accurate in the minority classes compared to $\mathcal{E}_3$, $\mathcal{E}_1$'s scales are still larger than $\mathcal{E}_3$'s. This explains why ACE without $L_{com}$ has the best performance in the head categories. In the experiments on SSC, where the experts learn to classify $\mathcal{C}_i$ but cannot distinguish the untrained categories, resulting in the obvious dominance of the expert trained with the full set in all splits, making other experts useless. 

Results here are inspiring: different from most exiting works that try to eliminate the bias, we utilize it. The data re-balancing is embedded in the data assignment to ensure more exposure of the minority. The individual back-propagation of each expert will not hurt the representation learning. Therefore, $L_{com}$ decouples the representation learning and classifier training in one stage.

\begin{table}[h]
\begin{center}
\small
\begin{tabular}{c|c|c c c}
\toprule[1.5pt]
\textbf{Methods} &  \textbf{All} & \textbf{Many} & \textbf{Medium} & \textbf{Few}\\\hline\hline
 \textbf{ACE (With $L_{com}$)} &  \textbf{49.4} & \textbf{66.1} & \textbf{55.7} & \textbf{23.5}\\
 \rowcolor{gray!10} Expert 1 & 41.9  & 71.2 & 40.2 & 10.7\\
 \rowcolor{gray!10} Expert 2 &  30.7 & 19.9 & 53.7 & 17.7\\
 \rowcolor{gray!10} Expert 3 &  21.8 & 0.0 & 38.7 & 27.8\\\hline
\textbf{Without $L_{com}$} & \textbf{47.2}& \textbf{71.5} & \textbf{49.4} &\textbf{17.5}\\
 \rowcolor{gray!10} Expert 1 &  42.0 & 71.0 & 40.9 & 10.5 \\
 \rowcolor{gray!10} Expert 2 &  31.1 & 19.4 & 53.8 & 19.4\\
 \rowcolor{gray!10} Expert 3 &  22.0 & 0.0 & 38.8 & 28.3\\
 \textbf{With SSC} & \textbf{43.4} & \textbf{65.1} & \textbf{44.4} & \textbf{18.0}\\
 \rowcolor{gray!10} Expert 1 &  41.6 & 68.2 & 41.2 & 12.1\\
 \rowcolor{gray!10} Expert 2 &  16.0 & 2.4 & 26.5 & 19.9\\
 \rowcolor{gray!10} Expert 3 &  21.4 &  0.0 & 38.6 & 26.7\\
\bottomrule[1.5pt]
\end{tabular}
\end{center}
\caption{Overall and many-/medium-/few-shot split top-1 accuracy on CIFAR100-LT-100 of the three model. The results are consistent with our analysis that without complementary loss, the experts are competing, so the results tend to average. Split-specific classifiers (SSC) depends mostly on $\mathcal{E}_1$.}
\label{tab:abs-lcom}
\end{table}

\textbf{Effectiveness of distribution-aware optimizer.} The distribution-aware optimizer controls the learning speed of each expert with various data assignments. In this section, we compare the linear scaling rule with the square-root scaling \cite{krizhevsky2014one} and a uniform optimizer. \cite{krizhevsky2014one} indicates when multiplying the batch size by $S$, one should multiply the learning rate by $\sqrt S$ to keep the variance in the gradient expectation constant. For a uniform optimizer, all the experts share the same $\eta$, \ie,
\begin{equation}
    \begin{aligned}
        \eta_i^{sqrt} = \eta_0 \cdot \sqrt {\frac{\sum_{c \in \mathcal{C}_i} {n_c}}{\sum^{\mathcal{C}} {n_j}}},
        \eta_i^{uni} = \eta_0
    \end{aligned}
\end{equation}

The training will be more sensitive to the variance of data with a larger learning rate. For the experts trained by minority splits, we have $\eta_i^{uni}>>\eta_i^{sqrt}>\eta_i^{linear}$. The comparison of the results is shown in Table~\ref{tab:abs-lr}. All three schemes produce better results than baseline. $\eta_i^{uni}$ promotes the higher improvements in the majority categories, while significantly decreases the tails. The reason is several experts converge too early and thus not effective due to over-fitting. $\eta_i^{sqrt}$ and $\eta_i^{linear}$ show similar performance, while $\eta_i^{linear}$ is better in medium and few-shot splits. By comparing $\eta_i^{sqrt}$ and $\eta_i^{linear}$, we observe learning rate is not the principal reason for accuracy booms. We conclude that selecting a proper optimization scheme with respect to the data distribution can effectively benefit the overall performance.
%\begin{figure}[h]
%\begin{center}
%\includegraphics[width=\linewidth]{LaTeX/Figures/ablation-lr.jpg}
%\end{center}
%   \caption{Comparisons of linear, square-root and uniform learning rate scaling schemes on CIFAR100-LT-100. }
%\label{fig:abl-lr}
%\end{figure}

\begin{table}[h]
\begin{center}
\small
\begin{tabular}{c|c|c c c}
\toprule[1.5pt]
\textbf{Scheme} &  \textbf{All} & \textbf{Many} & \textbf{Medium} & \textbf{Few}\\\hline\hline
 \textbf{Linear} &  49.4 & 66.1 & 55.7 & 23.5\\
\textbf{Square-root} &  49.1 & 67.1 & 55.2 & 22.1 \\
\textbf{Uniform}   & 41.7 &  69.7 & 39.9 & 10.7\\
\bottomrule[1.5pt]
\end{tabular}
\end{center}
\caption{Comparisons of learning rate scaling schemes on CIFAR100-LT-100.}
\label{tab:abs-lr}
\end{table}

\begin{figure}
\begin{center}
\includegraphics[width=0.9\linewidth]{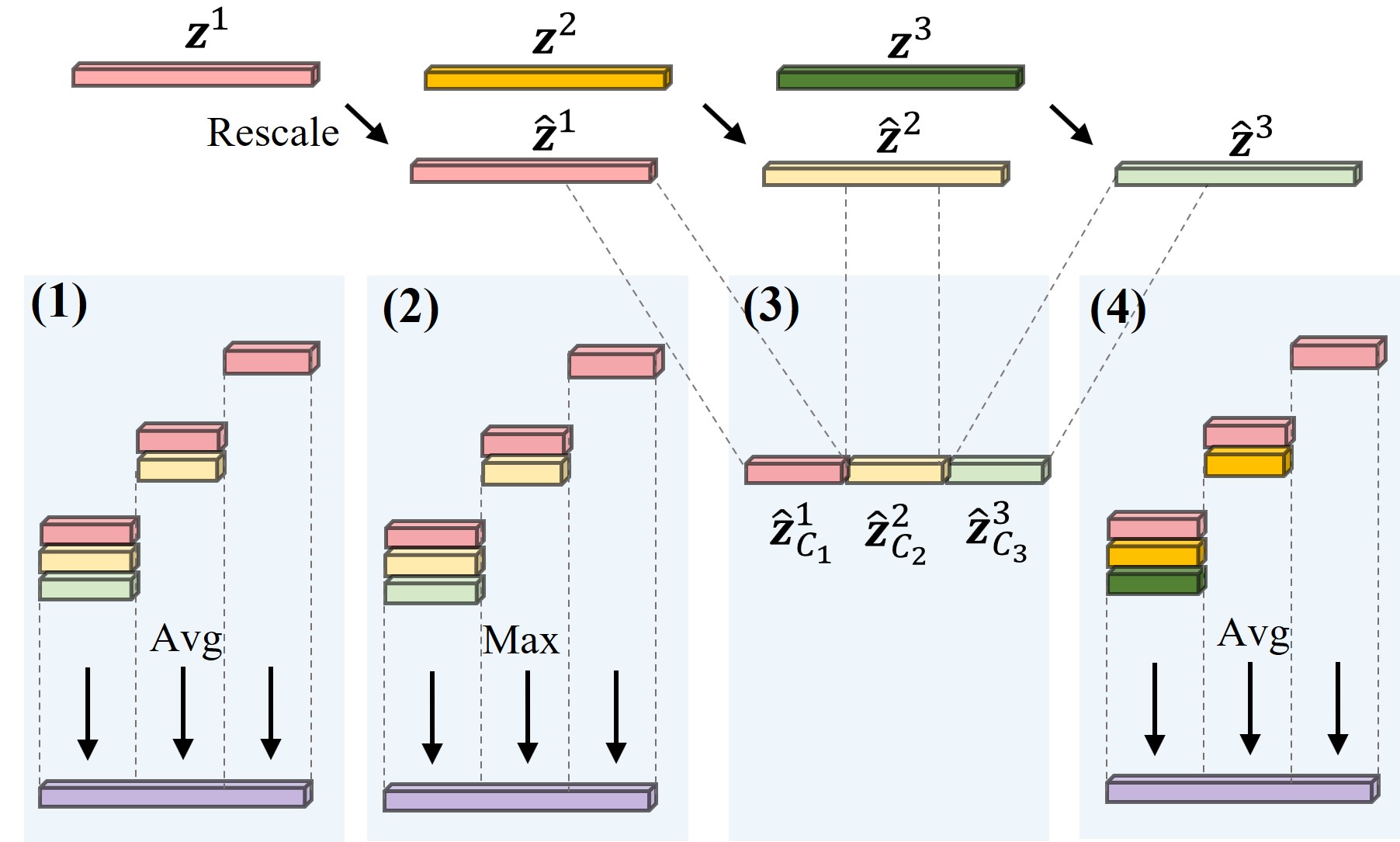}
\end{center}
   \caption{Illustration of variants of output aggregation methods.}
\label{fig:abl-out}
\end{figure}

\textbf{Effectiveness of group average output aggregation.} We compare different aggregation methods of the output logits $\{\textbf{z}^{i}\}$ from $K$ experts. Four variants of the aggregation methods are shown in Figure~\ref{fig:abl-out} (3 experts), the formulations are as follows,
\begin{itemize}
    \item (ACE) $\textbf{o} = 1 / |\{ \hat{\textbf{z}}^{i}_{\mathcal{C}_i}\}| \sum \{ \hat{\textbf{z}}^{i}_{\mathcal{C}_i}\}$,
    \item (2): Group Max, $\textbf{o}=\max \{ \hat{\textbf{z}}^{i}_{\mathcal{C}_i}\}$,
    \item (3): Group Concat, $\textbf{o}=\textbf{z}^{i}_{c \in \mathcal{C}_i,c \not \in \mathcal{C}_j, i\neq j}$,
    \item (4): Group Average without scaling,\\ $\textbf{o}=1 / |\{ \textbf{z}^{i}_{\mathcal{C}_i}\}| \sum \{ \textbf{z}^{i}_{\mathcal{C}_i}\}$.
\end{itemize}

Comparisons between (ACE) and (4) in Table~\ref{tab:abs-agg} shows that the design of scaling is for preserving the accuracy of the head classes. Computing the maximum by groups over the scaled logits (2) also suppresses the performance on the heads, as the experts for small classes are easier to overfit and thus overconfident. Concatenating the result of each dominating group of the experts amplifies the drawbacks of overconfidence, and experts competes each other. Overall, merging multiple experts is a trade-off for one-stage methods, in which all experts are trained from scratch. On the other hand, our ACE balances them by adjusting learning speed and with complementary loss, achieving improvements for all groups. 

\begin{table}[h]
\begin{center}
\small
\begin{tabular}{c|c|c c c}
\toprule[1.5pt]
\textbf{Aggregation} &  \textbf{All} & \textbf{Many} & \textbf{Medium} & \textbf{Few}\\\hline\hline
Group Avg w/ scaling (ACE) &  49.4 & 66.1 & 55.7 & 23.5\\
Group Max (2) &  43.4 & 47.5 & 54.2 & 26.5 \\
Group Concat (3)   & 37.7 &  30.3 & 50.2 & 22.9 \\
Group Avg w/o scaling (4)  & 46.7 & 49.5  &53.0 & 36.5\\
\bottomrule[1.5pt]
\end{tabular}
\end{center}
\caption{Ablation study on aggregation of the outputs on CIFAR100-LT-100.}
\label{tab:abs-agg}
\end{table}

\textbf{Effectiveness of data augmentation.} Mixup data augmentation is shown useful in previous studies \cite{zhou2020bbn, chou2020remix, zhang2021bag}. The results with different Mixup hyper-parameters $\alpha$ are shown in Table\~ref{tab:mixup}, which shows $\alpha$ will help to learn more generalizable and robust representations but does not have a significant impact on the overall performance.

\begin{table}[t]
\begin{center}
\small
\begin{tabular}{p{0.8cm}|c|c c c}
\toprule[1.5pt]
\textbf{$\alpha$} &  \textbf{All} & \textbf{Many} & \textbf{Medium} & \textbf{Few}\\\hline\hline
0 &  48.6 & 60.4 & 54.9 & 28.4\\
0.1 &  48.7 & 63.3 & 53.1 & 27.3 \\
0.2 & 48.1 &  65.7 & 53.1 & 22.9 \\
\textbf{0.3} & \textbf{49.4} & \textbf{66.1} & \textbf{55.7} & \textbf{23.5}\\
0.4 & 49.0 & 64.5  & 53.9 & 27.2\\
\bottomrule[1.5pt]
\end{tabular}
\end{center}
\caption{Ablation study on the Mixup parameter on CIFAR100-LT-100.}
\label{tab:abs-agg}
\end{table}

\section{Conclusion}
In this paper, extensive experiments on existing long-tailed recognition algorithms reveal the contradiction between biased representation learning and unbiased classifier learning. We proposed a multi-expert network that optimizes the two in a uniform network. Complementary constraints in data and objective function are applied to suppress the effects of non-targeted groups and promote both of the dominating and minority groups. Besides, a distribution-adaptive optimization scheme helps to adjust the learning paces of each expert to avoid over-fitting. ACE becomes the new SOTA among all one-stage long-tailed recognition methods with 3$\sim$10$\%$ accuracy gain, and is the first one that improves performance on all three frequency splits. With the equivalent strong performance to the multi-stage methods, there is great potential to extend well-formulated one-stage ACE to complex computer vision tasks like detection and segmentation.
{\small \balance
\bibliographystyle{ieee_fullname}
\bibliography{egbib}
}

\end{document}